# Enhanced Penalty-based Bidirectional Reinforcement Learning Algorithms

Sai Gana Sandeep Pula[1], Sathish A. P. Kumar[1], Sumit Jha[2] and Arvind Ramanathan[3]
[1]Department of Computer Science, Cleveland State University, Cleveland, OH USA
[2]School of Computing and Information Sciences, *Florida International University, Miami, FL, USA*
[3]Data Science and Learning Division, *Argonne National Laboratory , Lemont, IL, USA*
s.pula19@vikes.csuohio.edu; s.kumar13@csuohio.edu; sjha@fiu.edu; ramanathana@anl.gov

*Abstract*— This research focuses on enhancing reinforcement learning (RL) algorithms by integrating penalty functions to guide agents in avoiding unwanted actions while optimizing rewards. The goal is to improve the learning process by ensuring that agents learn not only suitable actions but also which actions to avoid. Additionally, we reintroduce a bidirectional learning approach that enables agents to learn from both initial and terminal states, thereby improving speed and robustness in complex environments. Our proposed Penalty-Based Bidirectional methodology is tested against Mani skill benchmark environments, demonstrating an optimality improvement of success rate of approximately 4% compared to existing RL implementations. The findings indicate that this integrated strategy enhances policy learning, adaptability, and overall performance in challenging scenarios.

## I. INTRODUCTION

### A. Motivation

The increasing complexity of real-world environments necessitates the development of reinforcement learning (RL) algorithms that not only optimize rewards but also prioritize robustness and safety. Traditional RL frameworks often fail to adequately penalize undesirable actions, resulting in suboptimal decision-making and unsafe trajectories. Addressing these shortcomings is essential for applications in robotics, autonomous systems, and dynamic task environments where reliability is critical. This research is driven by the need to enhance RL's adaptability and learning efficiency through an innovative bidirectional penalty-based approach. By incorporating penalties for invalid actions and leveraging bidirectional learning, agents can gain a deeper understanding of task dynamics while avoiding violations of constraints. The integration of forward and reverse learning processes creates a comprehensive mechanism for exploring optimal paths, thereby improving success rates and fostering effective policy development. This approach aims to establish new benchmarks in RL performance, particularly in tackling complex, high-dimensional challenges.

### B. Problem Statement

Reinforcement learning (RL) algorithms often face significant challenges in complex environments due to inefficient exploration and unsafe decision-making. Traditional RL approaches tend to prioritize maximizing rewards without adequately considering the repercussions of undesirable actions, which results in suboptimal trajectories and limited adaptability. Furthermore, existing methods frequently lack the mechanisms needed to simultaneously utilize information from both initial and goal states, impeding learning efficiency in high-dimensional tasks. The absence of a comprehensive strategy to balance exploration with safety poses obstacles to achieving robust and reliable policies. This research seeks to address these shortcomings by introducing a bidirectional penalty-based framework, thereby enhancing learning efficiency and ensuring adherence to constraints in dynamic environments.

These algorithms face notable limitations in complex environments, such as the pick cube task in the man skill benchmark. Many, like Proximal Policy Optimization (PPO), require extensive data to perform reliably, making them less suited for real-world applications. Algorithms such as Soft Actor-Critic (SAC) and TD-MPC2 demand substantial computational resources due to complex models and hyperparameter tuning. Traditional methods struggle with sparse rewards, while approaches like Reinforcement Learning with Policy Distillation (RLPD) need careful adjustments to avoid policy degradation. Additionally, current methods like Diffusion Policy may have difficulties in high-dimensional action spaces, where precise control is necessary to prevent errors.

| Algorithm | Limitations | Justification |
|---|---|---|
| PPO | Unsafe trajectories due to lack of penalization for invalid actions. | Penalty functions guide agents to avoid undesirable actions, ensuring safer and more efficient learning. |
| RLPD | Inefficient in sparse-reward environments policy degradation during distillation. | Penalty-based learning reinforces adherence to productive actions, mitigating degradation and improving stability. |
| SAC | High computational cost; challenges balancing exploration and exploitation. | Penalties reduce unproductive exploration, while bidirectional learning accelerates convergence. |
| TD-MPC-2 | Limited to unidirectional learning computationally intensive. | Bidirectional approaches leverage both forward and reverse trajectories, optimizing learning efficiency |
| Diffusion Policy | Struggles with high-dimensional spaces prone to errors. | Bidirectional penalties improve adaptability and robustness in dynamic, high-dimensional tasks. |

Fig. 1. RL Algorithm, it's limitations and Bidirectional Penalty based justification.

By Observing all these limitations of existing algorithms, our approach Penalty based Bidirectional perception, equipping robots to adapt, learn, and make decisions in ever-changing environments.

## C. Traditional Reinforcement Learning

Reinforcement Learning (RL) algorithms are essential in robotic perception, equipping robots to adapt, learn, and make decisions in ever-changing environments. These algorithms significantly improve object recognition, tracking, and manipulation by allowing robots to enhance their accuracy and precision through feedback, which is critical for tasks like picking and handling objects. Reinforcement learning is a computational framework focused on the determination of optimal actions in various situations to maximize a cumulative reward signal. It involves learning a policy that maps states of the environment to actions, thereby enabling an agent to make sequential decisions aimed at achieving the highest possible reward over time. [15]. Moreover, RL is vital for path planning and navigation, enabling robots to dynamically alter their routes to circumvent obstacles.

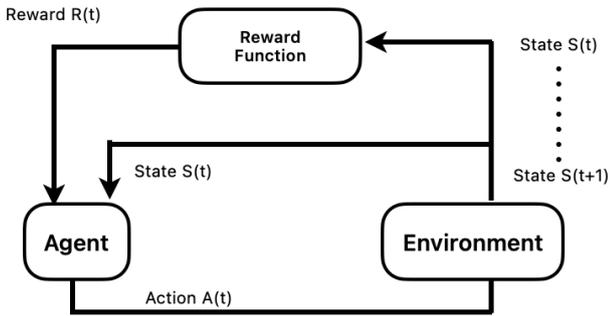

Fig. 2. A simple architecture of how Reinforcement Learning works.

When the agent is in state S(t), it perceives its environment and strives to adapt to it. To facilitate this adaptation, the agent employs a program function known as the Reward function. This function assists the agent in monitoring all valid actions A(t) that can be taken to transition to the next state with a reward value R(t) for a specific trajectory.

## D. Penalty Functions

Incorporating penalty functions into reward structures within reinforcement learning (RL) frameworks significantly enhances agent decision-making processes. By integrating penalties, learning agents are guided to avoid suboptimal actions and misjudgments while simultaneously enabling them to concentrate on maximizing rewards. This methodological approach strikes a critical balance between exploration and adherence to established constraints, thereby fostering the development of robust and optimal policies. Consequently, this framework promotes the learning algorithms' effectiveness and enhances overall agent performance in complex environments.

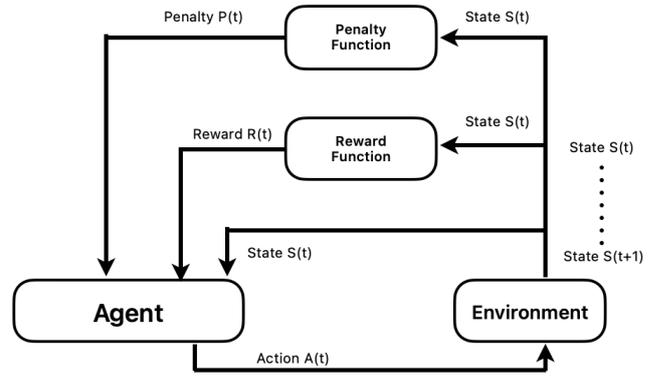

Fig. 3. Penalty incorporated RL algorithm.

Recent advances in bilevel optimization have expanded its application to RL, addressing complex decision-making issues where incentives and objectives may be misaligned. In bilevel RL, The lower-level problem in bilevel reinforcement learning (RL) transitions from the realm of smooth optimization to the domain of policy optimization within the context of reinforcement learning. allowing for a dynamic interplay between penalty functions and reward structures in complex environments [1]. Implementing penalties within learning algorithms enables agents to mitigate maladaptive learning behaviors, facilitating more efficient and reliable navigation through complex environments. This approach fosters a more robust decision-making framework, allowing for the practical identification and avoidance of suboptimal strategies.

## E. Why do we need Penalty functions?

In Reinforcement Learning (RL), the no-reward concept is often employed to discourage undesirable actions implicitly by withholding positive reinforcement. However, this approach raises a critical question: Why should we explicitly use penalty functions when no-reward achieves a similar effect? The answer lies in the nuanced impact of penalty enforcement versus passive discouragement, particularly in high-risk and safety-critical environments.

To illustrate this distinction, consider the case of a self-driving car navigating an urban environment. The primary objective of the vehicle is to reach its destination efficiently while adhering to traffic regulations. A conventional reward-based system may allocate +10 points for reaching the destination and +5 points for stopping at a red light. However, if the system relies solely on a no-reward strategy for undesirable behaviors, such as running a red light, the car may fail to recognize the inherent risk associated with this action. Since no immediate negative consequence is assigned, the model may still attempt the behavior if it identifies a shortcut or optimization strategy that bypasses traffic rules.

Conversely, by introducing an explicit penalty function—for instance, applying -20 points for running a red light—the system actively discourages such behavior. This distinction is crucial because a no-reward strategy merely fails to reinforce an action, while a penalty function directly penalizes unsafe decisions, leading to more robust policy learning. In safety-critical applications such as autonomous driving, financial trading, and healthcare, where incorrect decisions can have severe consequences, the use of penalty functions becomes imperative. A no-reward approach may allow models to explore suboptimal or risky actions, whereas explicit penalties create stronger constraints on decision-making, ensuring compliance with safety regulations and ethical considerations.

Therefore, explicitly incorporating penalty functions in RL models enhances learning efficiency, promotes safer decision-making, and reduces the likelihood of policy exploitation in high-risk environments. By reinforcing the avoidance of negative behaviors rather than merely omitting rewards, RL systems achieve greater reliability, making them more suitable for real-world deployment.

### F. Bidirectional Approach

Additionally, bidirectional learning in RL combines forward and reverse approaches, allowing agents to learn from the initial and final stages of tasks. This method accelerates learning and improves robustness by providing a more comprehensive understanding of the environment and task dynamics. Reverse-forward curriculum learning provides a robust framework for enhancing agent exploration and problem-solving capabilities in complex tasks. This approach facilitates an iterative process that alternates between backward and forward progressions, thereby optimizing the utility of limited demonstrations and promoting accelerated learning outcomes. By leveraging this dual progression strategy, practitioners can significantly improve the efficiency and effectiveness of training paradigms in various domains. [2]. Combining these methods aims to improve the speed, efficiency, and reliability of agents learning in complex scenarios and facilitate the development of more adaptive and optimal policies.

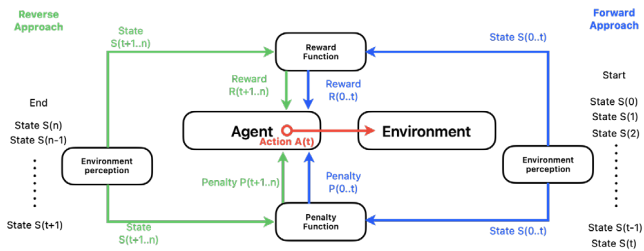

Fig. 4. Bidirectional Penalty based RL algorithm.

### G. Comparision Between RFCL and Bidirectional Approach

The Reverse Forward Curriculum Learning (RFCL) method provides a structured mechanism to enhance reinforcement learning through a curriculum-based training approach. Inspired by RFCL, our approach introduces a bidirectional method that integrates both forward and reverse learning techniques while incorporating penalty functions to refine learning trajectories and improve agent decision-making.

Although existing algorithms have shown success in structured learning and demonstrated individual strengths in achieving higher success rates, they are limited by their reliance solely on forward learning. This observation raised a pivotal question: What if we allow learning to progress from both ends, thereby refining policy optimization by leveraging both initial and terminal states?

By utilizing the structured progression of curriculum learning, this research focuses on introducing a bidirectional approach to existing reinforcement learning algorithms. This includes the integration of both forward and reverse curriculum learning alongside the application of penalty functions to optimize decision-making.

When comparing performance metrics, our bidirectional approach consistently demonstrated greater adaptability and robustness than RFCL. In our experimental evaluations using the Mani Skill benchmark, we observed an approximately 4% increase in success rates over traditional reinforcement learning (RL) implementations, which allowed us to outperform RFCL in complex, long-horizon tasks. The incorporation of penalty functions in our method further enhanced the stability of the agent by discouraging unsafe actions and promoting optimal trajectories—an aspect that RFCL does not explicitly address. While RFCL excels in staged, curriculum-based reinforcement learning, our approach refines and expands on its foundational principles by integrating these penalty functions. This ensures the learning process remains stable, efficient, and adaptable in various scenarios. This comparison illustrates that we have developed a more resilient and generalizable reinforcement learning framework by drawing inspiration from RFCL and extending it with bidirectional, penalty-based learning from RFCL. We have developed a more resilient and generalizable reinforcement learning framework by extending it with bidirectional, penalty-based learning.

### H. Our Contribution

This work proposes a novel reinforcement learning (RL) framework that integrates penalty-based optimization with a bidirectional learning approach, addressing critical limitations in existing methodologies. Specifically, this research makes the following contributions:

- Combines penalty functions with bidirectional learning to create a unified framework applicable to multiple RL algorithms, including Proximal Policy Optimization (PPO), Reinforcement Learning with Policy Distillation (RLPD), and Soft Actor-Critic (SAC). This dual enhancement ensures safer trajectories and faster convergence compared to standalone methods.

- The penalty mechanism discourages unsafe and inefficient actions while the bidirectional learning approach leverages information from both initial and goal states, significantly improving learning speed, stability, and adaptability in high-dimensional and dynamic environments.

- Unlike prior works that focus on single techniques or limited benchmarks, this approach is evaluated across a diverse set of RL tasks in the Mani Skill environment. Experimental results show consistent improvements in success rates, with up to a 4% increase in performance compared to baseline algorithms.

- Introduces a modular enhancement that can seamlessly integrate into existing RL frameworks with minimal modifications. This feature makes the approach broadly applicable to real-world problems, such as robotic manipulation and autonomous navigation, where safety and efficiency are paramount.

- Provides a rigorous mathematical foundation for the proposed penalty functions and bidirectional approach, along with extensive experimental validation using quantitative metrics like success rates, convergence speed, and policy robustness.

Previous studies have demonstrated the separate implementations of penalty functions and bidirectional approaches [1][2]. By addressing gaps in reward optimization, policy stability, and learning efficiency, this research bridges the limitations of prior work, establishing a benchmark for robust and adaptable RL systems in complex environments. However, these methods have primarily served as introductory frameworks and have been limited in their application to specific research contexts. Building upon the foundational ideas presented by these researchers, this study aims to integrate both approaches, thereby introducing a novel methodology for their incorporation into established algorithms.

## II. LITERATURE REVIEW

Recent reinforcement learning (RL) advancements have focused on tackling critical challenges such as sample efficiency, policy robustness, and adaptability within complex environments. Core techniques, including bilevel optimization, bidirectional learning, and model-based RL, drive this evolving landscape in ways that significantly refine policy development processes.

### A. Background

Several algorithms serve as baselines for addressing tasks in the Mani Skill Benchmark, including PPO, Diffusion Policy, RLPD, TDMPC-2, SAC, RFCL, and Behavior Cloning. However, none of these implementations utilize a penalty function, which is the primary focus of our contribution. Furthermore, we believe that combining a bidirectional approach with penalty functions can significantly enhance the success rates of implementations within the Mani Skill Benchmark.

The ManiSkill3 benchmark, an advancement over the original SAPIEN ManiSkill, addresses gaps in The research aims to enhance generalizable manipulation skills through the provision of a comprehensive suite comprising 20 distinct families of manipulation tasks. This is supported by an extensive collection of over 2,000 object models and more than 4 million demonstration frames.[3] to enhance research reproducibility and policy evaluation in diverse tasks. *Gu et al.* also noted that ManiSkill2's unified interface facilitates the rapid development of algorithms for visual input learning. allowing agents to achieve high processing rates for complex manipulation tasks. This benchmark has facilitated significant progress in evaluating RL agent adaptability across varied robotic manipulation tasks. The outcomes of findings from the maniskill benchmark are presented in reference [9].

*Hansen et al.'s* TD-MPC2 algorithm exemplifies the leap in scalable, model-based RL, particularly in handling large datasets and multi-task settings without extensive tuning. The team stated that TD-MPC2 study focuses on the development of generalist world models that are designed to learn from extensive, uncurated datasets encompassing diverse task domains, embodiments, and action spaces. The aim is to enhance the versatility and applicability of these models across a broad spectrum of scenarios. addressing the scalability challenge in continuous control tasks [4]. This innovation allows TD-MPC2 to outperform baseline methods across 104 distinct tasks, demonstrating its effectiveness in complex control settings.

*Chi et al.'s* Diffusion Policy leverages diffusion models for visuomotor policy learning, which is especially valuable in robotic tasks requiring high precision. The authors remarked that the Diffusion Policy framework demonstrates the capacity to represent a wide variety of normalizable probability distributions, thereby facilitating the handling of high-dimensional action sequences. This capability underscores its versatility and adaptability in complex decision-making scenarios. which is essential for tasks with complex, multimodal action distributions [5]. This approach achieves remarkable stability during training by conditioning on visual inputs, which facilitates efficient policy learning and enhances performance across diverse manipulation benchmarks.

Clean RL, introduced by *Huang et al.,* streamlines RL experimentation by offering concise, monolithic implementations of Deep Reinforcement Learning (DRL) algorithms. allowing researchers which is essential to comprehend the comprehensive implementation details to effectively and expediently prototype innovative features.[6]. This approach improves the reproducibility and transparency of RL research, providing a platform for researchers to explore RL techniques with a simplified coding structure that facilitates both learning and innovation.

### B. Related Works

The motivation for this study arises from the necessity to introduce a novel approach that can seamlessly integrate with existing algorithms, thereby augmenting their effectiveness and improving success rates.

*Bilevel Optimization and Penalty-Based Approaches:*

*Shen et al.* proposed a penalty-based approach for bilevel RL, designed to tackle hierarchical control tasks like Stackelberg games and reinforcement learning from human feedback (RLHF). They highlighted, The principal technical challenge associated with bilevel optimization resides in effectively managing the constraints imposed by the lower-level problem. This complexity necessitates a nuanced approach to ensure that the interdependencies between the two levels of optimization are rigorously addressed. [1] which in bilevel RL requires overcoming the non-convex nature of the discounted reward function. Their framework introduces value and Bellman penalties, which In order to formalize the optimality condition associated with the lower-level reinforcement learning (RL) problem, it is essential to rigorously characterize the criteria under which the solution achieves equilibrium or optimal performance. This involves identifying the parameters and constraints that govern the decision-making process, thereby ensuring that the derived policies conform to established standards of optimality within the specified environment. Through this analytical framework, one can effectively elucidate the interdependencies of state transitions, actions, and rewards, ultimately contributing to a comprehensive understanding of the dynamics at play in the lower-level RL paradigm.[1] effectively guiding policy decisions while ensuring stability.

*Curriculum Learning for Sample Efficiency:*

The Reverse Forward Curriculum Learning (RFCL) method introduced by *Tao et al.* enhances demonstration and sample efficiency through a unique curriculum structure. RFCL In the work of Tao et al., the authors present a comprehensive analysis that contributes significantly to the existing body of literature. Their findings elucidate key concepts and offer valuable insights into the subject matter, thereby enhancing our understanding of the topic at hand as the agent first learns to solve tasks in constrained environments before gradually generalizing to broader scenarios [3]. This method has been particularly effective for high-precision and long-horizon tasks, often solving tasks that traditional approaches struggle with due to sparse rewards or high-dimensional state spaces.

| Algorithm | Key strengths | Limitations | How our approach address this limitations |
|---|---|---|---|
| PPO | Balances learning efficiency with stability, reducing the risk of disruptive policy updates. | Unsafe trajectories due to lack of penalization for invalid actions. | Penalty functions guide agents to avoid undesirable actions, ensuring safer and more efficient learning. |
| RLPD | Enhances learning efficiency by utilizing offline data, reducing the need for extensive online interactions. | Inefficient in sparse-reward environments; policy degradation during distillation. | Penalty-based learning reinforces adherence to productive actions, mitigating degradation and improving stability. |
| SAC | Promotes broad exploration due to entropy maximization, improving stability and robustness. And deals with soft bodies like liquid containers. | High computational cost; challenges balancing exploration and exploitation. | Penalties reduce unproductive exploration, while bidirectional learning accelerates convergence. |
| TDMPC-2 | Handles large datasets and multi-task settings with minimal tuning. Efficiently scales to complex, high-dimensional control tasks. | Limited to unidirectional learning; computationally intensive. | Bidirectional approaches leverage both forward and reverse trajectories, optimizing learning efficiency. |
| Diffusion Policy | Enhances performance in visuomotor tasks with complex action distributions. Allows smooth, stable policy learning for complex actions. | Struggles with high-dimensional spaces; prone to errors. | Bidirectional penalties improve adaptability and robustness in dynamic, high-dimensional tasks. |

This table summarizes key limitations of widely used RL algorithms (e.g., PPO, RLPD, SAC) and explains how bidirectional penalty-based approaches address these challenges. It highlights issues such as unsafe trajectories, sparse rewards, and high-dimensional action spaces, providing justification for the proposed enhancements.

The Penalty-Based Bidirectional Reinforcement Learning (RL) framework distinguishes itself from existing approaches by effectively integrating penalties with bidirectional learning. This dual strategy not only addresses the inherent limitations of individual algorithms but also combines their strengths, resulting in improved safety, efficiency, and adaptability. In contrast to traditional RL methods, which typically perform well in isolated scenarios, the proposed framework showcases versatility and robustness across a range of RL tasks, as demonstrated by its performance on the ManiSkill benchmark. Its ability to seamlessly integrate with various algorithms with minimal adjustments makes this framework a practical solution for real-world challenges, including robotic manipulation and autonomous navigation.

## III. METHODOLOGY

### A. Generic Example of Proposed Solution:

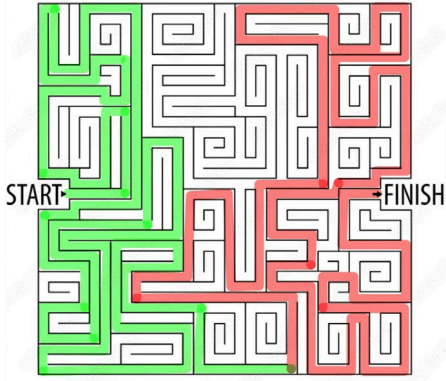

Fig. 5. An example of bidirectional learning, highlighting its reciprocal nature in the learning process.

The maze diagram effectively illustrates the working of a bidirectional penalty-based algorithm. In this approach, perception begins from both the start and finish points of the maze, exploring potential routes simultaneously from each end. As the algorithm progresses, it assigns rewards for each of the correct decisions and weighted penalties for incorrect perception to each path segment based on obstacles or the complexity of navigation. The highlighted paths from the start (green) and finish (red) visually represent these explorations. As the algorithm advances, the paths converge or come close to each other in certain sections, where it evaluates these meeting points to identify the most efficient route based on cumulative penalties. By continually updating based on penalty values, the algorithm dynamically selects the least costly path, even if it requires adjusting routes midway, as demonstrated where the green and red paths approach but do not overlap. This process ensures an optimized path selection by balancing between the shortest route and the lowest penalty, showcasing the bidirectional nature of the algorithm.

In this methodology, we introduce a novel bidirectional, penalty-based algorithm for guiding agents through complex maze environments. By leveraging both a reverse and forward curriculum, this approach enables efficient exploration and learning, overcoming the limitations often found in traditional reinforcement learning (RL) frameworks. This method begins with the reverse curriculum, which initializes the agent near goal states and gradually exposes it to more challenging start positions. This strategy efficiently addresses exploration bottlenecks, a common challenge in sparse-reward settings, as the agent can first achieve success in a narrower scope before expanding its capabilities. As noted, The reverse curriculum facilitates sample-efficient learning by drawing from a limited set of initial states present in the demonstrations. [2]

Following the reverse curriculum, a forward curriculum is employed to progressively increase the complexity of the agent's initial states. By training the agent on states closer to its current capacity, the forward curriculum ensures continuous but manageable learning progression. This structured transition allows the agent to generalize its policy across a broader distribution of initial states. It has been shown that combining both curriculum stages leads to greater demonstration and sample efficiency than applying either in isolation [2]. To incorporate these strategies in RL environments with bilevel structures, a penalty-based approach is adapted, where penalties reinforce desirable behaviors in lower-level RL problems. The penalty functions such as value and Bellman penalties are critical for shaping the learning trajectory and promoting convergence in bilevel optimization tasks, as they enable the system to adhere to both upper-level and lower-level objectives [1].

### B. Mathematical derivation

In this research, we propose a penalty-based approach to enhance reinforcement learning by incorporating penalties that discourage undesirable actions while guiding the agent toward optimal behavior. The penalty function consists of two main components: Deviation-based Penalty and Un-usage Action Penalty. The Deviation-based Penalty Function is defined as The value penalty function serves as a quantifiable measure of the optimality gap at the lower level, thereby ensuring that the agent's policy is in close alignment with the intended outcomes. This is typically achieved through the application of metrics such as the Euclidean distance, which evaluates the disparity between optimal and actual trajectories. [1] these structure defines and tracks optimal trajectories for both forward and reverse directions, in which the agent's actions are compared.

$$P_{deviation}(\tau_c) = ||\tau_{optimal} - \tau_{actual}||^2$$

We $\tau_{optimal}$, represent optimal trajectory, and $\tau_{actual}$, represents the agent's actual trajectory. The penalty is calculated based on the Euclidean distance between the optimal and actual trajectories, ensuring that the agent minimizes deviations from the ideal path.

The unusage Action Penalty is intended to penalize actions that are left unused, helping the agent avoid undesirable actions. It is defined as RFCL implements a system of penalties for actions that remain unutilized, aimed at enhancing agent behavior. These penalties are formulated to highlight and incentivize actions that are closely aligned with the trajectory deemed necessary for achieving successful outcomes.[2]

$$P(\tau_c) = Unused\ (\tau_c)\ X\ C_{unused}$$

where Unused($\tau_c$) is a binary indicator (1 if the action is unused, 0 otherwise), and $C_{unused}$ is a constant penalty value applied to unused actions.

To determine the overall penalty, the Combined Penalty Function is expressed as The penalty function, integrated with balancing factors, is employed in conjunction with both reverse and forward learning paradigms. This approach facilitates agents in harnessing the advantages of learning from both past experiences and future predictions, thereby enhancing their overall adaptive capabilities in dynamically changing environments. [1]

$$P(\tau_c) = \omega_{deviation}\ X\ P_{deviation}(\tau_c) \\ + \omega_{unused}\ X\ P_{unused}(\tau_c)$$

where $\omega_{deviation}$ and $\omega_{unused}$ are weighting factors that balance the influence of the two penalty terms, these weights allow flexibility in controlling the importance of each type of penalty in the overall learning process.

The Mathematical Objective of the agent is formulated as follows:

$$\delta(\theta) = E_{\tau_c \sim \pi_\theta} \left[ \sum_t (R(c_t, a_t) - \rho \cdot P(\tau_c) - \beta \cdot KL(\pi_\theta(\tau|c_t) || \pi_{target}(\tau|c_t))) \right]$$

where:

- $R(c_t, a_t)$ is the reward function.
- $\rho$ is the penalty function weight, controlling the impact of penalties on learning.
- $P(\tau_c)$ is the combined penalty function.
- $\beta$ is the weight for the Kullback-Leibler (KL) divergence, which ensures the learned policy $\pi\theta$ remains close to the target policy $\pi_{target}$.

By combining these penalty functions with the reward structure, the agent can navigate complex environments more effectively, balancing the goals of constraint adherence and maximizing rewards. The integration of deviation and unused action penalties aims to foster more adaptive and optimal policy development, improving both the robustness and efficiency of learning in constraint-driven scenarios.

### C. Bidirectional Penalty Approach

This approach leverages both forward and reverse learning processes, thereby improving learning speed and robustness by considering the trajectory from both directions.

The bidirectional penalty function is comprised of several components:

The Forward Penalty approach ($P_f$) measures the Euclidean distance between the optimal and actual forward trajectories in the context of bilevel reinforcement learning, the lower-level problem is frequently restructured by employing a penalty function, which serves to impose costs on the transgressions of lower-level constraints. This approach allows for systematic enforcement of constraints while facilitating the optimization process at both levels of the hierarchical framework. [1]

$$P_F(\tau_F) = ||\tau_{optimal}, F - \tau_{actual}, F||^2$$

Here, ($\tau_{optimal}$,F) and ($\tau_{actual}$, F) represent the optimal and actual forward trajectories, respectively. This penalty measures the Euclidean distance between the optimal and actual forward trajectories to minimize deviations during forward learning. Reverse Deviation Penalty (Pr) similar to the forward penalty, measures the Euclidean distance between optimal and actual trajectories in reverse, encouraging alignment with desired outcomes during reverse learning [2]

$$P_R(\tau_R) = ||\tau_{optimal}, R - \tau_{actual}, R||^2$$

Similar to the forward penalty, this component measures the Euclidean distance between the optimal and actual trajectories but in the reverse direction, encouraging alignment with the desired outcome during reverse learning.

$$P_{unused,F}(\tau_F) = Unused(\tau_F) \times C_{unused}, F$$

If an action is unused in the forward trajectory, it is assigned a penalty defined by a constant ($C_{unused}$,F) determines how severely unused actions are penalized during the forward trajectory.

$$P_{unused,R}(\tau_R) = Unused(\tau_R) \times C_{unused}, R$$

Similarly, if an action is unused during the reverse trajectory, a penalty is assigned using the constant ($C_{unused}$,R). These constants help adjust the strength of the penalty, depending on the specific learning requirements or application environment. By setting appropriate values for these constants, the algorithm encourages agents to explore and utilize all possible actions, rather than ignoring certain actions that could lead to better outcomes. Larger values of these constants result in higher penalties, making unused actions more costly. If the forward learning phase is considered more critical for task success, ($C_{unused}$,F) can be set higher than ($C_{unused}$,F) or vice versa, depending on the domain or environment.

Combined Penalty Function for Bidirectional Learning

The combined penalty function ($P_{total}$) for bidirectional learning is formulated as the implementation of a bidirectional penalty function supplemented by balancing factors is employed in conjunction with both reverse and forward curriculum learning methodologies. This integrated approach ensures that the agent not only maximizes its rewards but also effectively mitigates constraint violations, thereby fostering the development of more robust and adaptable policies. [2]

$P_{total}(\tau) = \omega_F \times (P_F(\tau_F) + P_{unused,F}(\tau_F)) + \omega_R \times (P_R(\tau_R) + P_{unused,R}(\tau_R)))$

Where:

$\omega_F$ and $\omega_R$ are weights that balance the influence of forward and reverse penalties, allowing for flexibility in determining the contribution of each learning direction.

Here, The penalty function with balancing factors is used in collaboration with reverse forward curriculum learning, allowing agents to benefit from learning in both directions—forward and backward. By applying penalties in both directions and balancing their influence, this approach ensures that the agent not only maximizes rewards but also avoids constraint violations and learns more robust policies. The use of bidirectional penalties aims to enhance the speed, reliability, and adaptability of the agent's learning process, leading to improved performance in complex environments.

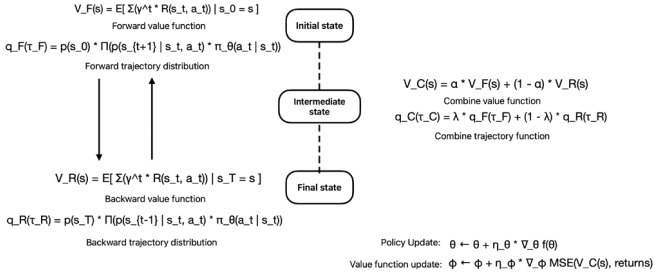

Fig. 6. The bidirectional learning framework that combines forward and backward value functions and trajectory distributions.

In this framework, $V_F(s)$ represents the forward value function, which is the expected cumulative reward starting from the initial state. $q_F(\tau_f)$ denotes the forward trajectory distribution, calculated based on the probability of transitioning forward from the initial state. $V_R(s)$ is the backward value function, indicating the expected reward when starting from the final state and moving backward. $Q_R(\tau_R)$ defines the backward trajectory distribution, capturing the probability of transitions when traced in reverse from the final state.

These components are combined to form the *combined value function* $V_C = \alpha \cdot V_F(s) + (1-\alpha) \cdot V_R(s)$ and combined trajectory function $q_c(\tau_c) = \lambda \cdot q_F(\mathcal{T}_F) + (1-\lambda) \cdot q_R(\tau_R)$ allowing the model to leverage information from both directions.

The policy update step $\theta \leftarrow \theta + n\theta \cdot \nabla\theta\, f(\theta)$ adjusts the agent's policy parameters, and the value function update $\Phi \leftarrow \Phi + n\Phi \cdot \nabla\Phi\, \text{MSE}(V_C(s), \text{returns})$ refines the combined value function to minimize mean squared error between predicted and actual returns. This approach enables improved learning efficiency by incorporating both forward and reverse dynamics into the value and policy updates.

*D. Numerical Example*

Consider a simplified reinforcement learning (RL) environment where an agent must navigate a 5x5 grid to reach the target at (5, 5) from its starting position at (1, 1). The task requires the agent to follow an optimal trajectory, utilizing the available actions (up, down, left, right, diagonal) to minimize deviations and reach the goal efficiently.

Optimal Trajectory: The ideal path ($\tau_{optimal}$) to the target is:
$(1,1) \rightarrow (2,2) \rightarrow (3,3) \rightarrow (4,4) \rightarrow (5,5)$
This trajectory is pre-determined and considered the shortest path while minimizing deviations.

Actual Trajectory However, the agent takes a slightly different path ($\tau_{actual}$):
$(1,1) \rightarrow (2,1) \rightarrow (3,2) \rightarrow (4,3) \rightarrow (5,5)$
Here, the agent deviates from the optimal trajectory at multiple points due to suboptimal decision-making.

*Step 1- Deviation Based Penalty Calculation:*

To evaluate the deviation of the agent's trajectory ($\tau_{actual}$) from the optimal trajectory ($\tau_{optimal}$), the **Euclidean distance** between corresponding points is calculated for each step.

Deviations at each step as follows.

From **(2, 1)** (actual) to **(2, 2)** (optimal):

$\|(2,1)-(2,2)\|_2 = (2-2)^2 + (1-2)^2 = 0+1 = 1$

From **(3, 2)** (actual) to **(3, 3)** (optimal):

$\|(3,2)-(3,3)\|_2 = (3-3)^2 + (2-3)^2 = 0+1 = 1$

From **(4, 3)** (actual) to **(4, 4)** (optimal):

$\|(4,3)-(4,4)\|_2 = (4-4)^2 + (3-4)^2 = 0+1 = 1$

Total deviation-based penalty.

Summing the deviations across all steps:

$P_{dev} = 1+1+1 = 3$

This penalty of **3** represents the total deviation from the optimal trajectory. Minimizing this penalty encourages the agent to stay closer to the optimal path.

*Step 2 – Unused Action Penalty:*

In this scenario, the agent ignored some valid actions that could have contributed to achieving the goal more effectively.

Unused Actions.

The agent did not use the "up" (↑) and "left" (←) actions during its trajectory.

A penalty is assigned for each unused action, with a constant penalty value $C_{unused} = 5$.

Unused action penalty calculation.

$P_{unused} = C_{unused} \cdot \text{Number of unused actions}$

$P_{unused} = 5 \cdot 2 = 10$

This penalty of 10 is added to discourage the agent from ignoring useful actions in future iterations.

*Step 3 – Combined Penalty*

The total penalty combines the deviation-based penalty ($P_{dev}$) and the unused action penalty ($P_{unused}$), using weights $\omega_{dev}$ and $\omega_{unused}$ to balance their importance.

Given weights,

$\omega_{dev} = 0.7$: The weight assigned to the deviation penalty.

$\omega_{dev} = 0.3$ The weight assigned to the unused action penalty.

Combined penalty calculation $P_{total}$

$P_{total} = \omega_{dev} \cdot P_{dev} + \omega_{unused} \cdot P_{unused}$

$P_{total} = 0.7 \cdot 3 + 0.3 \cdot 10$

$P_{total} = 2.1 + 3 = 5.1$

This total penalty of 5.1 is used to adjust the agent's reward for the trajectory.

### E. Pseudo Code

```
# Initialize environment and agent
initialize agent()
initialize environment()

# Set parameters for bidirectional penalty calculation
set penalty weights wF, wR   # Weights for forward and reverse penalties
set deviation_threshold      # Threshold for allowable deviation
set penalty_constant         # Constant penalty value for unused actions

# Training loop
for each episode in num_episodes:
    # Reset environment and agent state
    state = environment.reset()
    agent.reset()

    # Track actual and optimal trajectories
    optimal_forward_trajectory = calculate_optimal_trajectory(start=state, direction="forward")
    optimal_reverse_trajectory = calculate_optimal_trajectory(goal=environment.goal, direction="reverse")
    actual_forward_trajectory, actual_reverse_trajectory = [], []

    # Forward pass in the environment
    for each step in num_steps:
        # Agent takes action in forward direction
        action = agent.select_action(state)
        next_state, reward, done = environment.step(action)

        # Track the agent's forward trajectory
        actual_forward_trajectory.append(state)

        # Check if the action aligns with the optimal forward trajectory
        forward_deviation_penalty = calculate_deviation_penalty(actual_forward_trajectory, optimal_forward_trajectory, deviation_threshold)
        forward_unsafe_penalty = calculate_unused_action_penalty(action, penalty_constant, direction="forward")

        # Move to next state
        state = next_state
        if done:
            break

    # Reverse pass for backward trajectory
    reverse_state = environment.goal  # Start from goal for reverse trajectory
    for each reverse_step in num_steps:
        # Agent selects action for reverse path (e.g., inverse of forward actions)
        reverse_action = agent.select_reverse_action(reverse_state)

        next_reverse_state, reverse_reward, reverse_done = environment.reverse_step(reverse_action)

        # Track the agent's reverse trajectory
        actual_reverse_trajectory.append(reverse_state)

        # Calculate penalties for the reverse direction
        reverse_deviation_penalty = calculate_deviation_penalty(actual_reverse_trajectory, optimal_reverse_trajectory, deviation_threshold)
        reverse_unsafe_penalty = calculate_unused_action_penalty(reverse_action, penalty_constant, direction="reverse")

        # Move to the next reverse state
        reverse_state = next_reverse_state
        if reverse_done:
            break

    # Combined penalty function
    total_penalty = (wF * (forward_deviation_penalty + forward_unsafe_penalty)) + (wR * (reverse_deviation_penalty + reverse_unsafe_penalty))

    # Adjust rewards based on total penalties (example for reward modification)
    adjusted_reward = reward - total_penalty

    # Update agent with adjusted rewards
    agent.update_policy(adjusted_reward)
```

### IV. PENALTY CALCULATION

Deviation Penalty: Measures the Euclidean distance between the agent's actual trajectory and the optimal trajectory for both forward and reverse directions.

Unused Action Penalty: Penalty for actions not aligning with the optimal trajectory, calculated separately for forward and reverse paths.

Combined Penalty Function: Applies weighting factors $W_F$ and $W_R$ to the forward and reverse penalties, respectively, to obtain the total penalty.

Adjusted Reward: The agent's rewards are modified by subtracting the combined penalty, allowing it to learn from both directions, with penalties reinforcing optimal behavior.

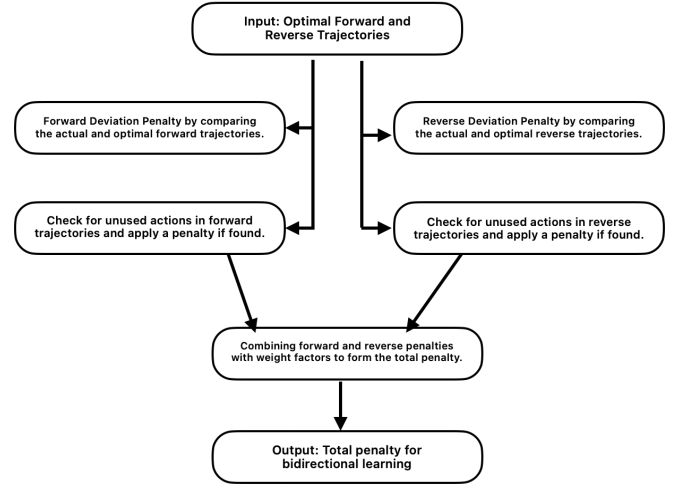

Fig. 7. The bidirectional approach to penalty calculation.

Thus highlighting the iterative processes involved in assessing penalties from multiple perspectives. This schematic representation facilitates a comprehensive understanding of the mechanisms at play in the evaluation and attribution of penalties within the outlined framework. The "Bidirectional Penalty Calculation Process" flowchart illustrates the systematic approach to assessing penalties within the bidirectional learning framework, as applied in reinforcement learning. This method leverages both forward and reverse trajectories, enabling the model to evaluate actions from multiple perspectives. The process begins with the "Optimal Forward and Reverse Trajectories," serving as a baseline against which actual trajectories are compared. As Shen, Yang, and Chen (2024) [1] discuss, such penalty-based approaches in bilevel reinforcement learning facilitate "guiding policy decisions while ensuring stability" by introducing deviation penalties for actions that stray from desired trajectories.

The penalty assessment includes two main components for each trajectory direction: the "Deviation Penalty," which measures the alignment between actual and optimal trajectories, and the "Unused Action Penalty," which applies to actions that do not contribute to the trajectory. These penalties are then combined with weight factors to form a "Total Penalty for Bidirectional Learning." This holistic penalty calculation method allows agents to balance the trade-off between optimal policy adherence and constraint violations, resulting in a more robust and adaptive policy framework. As noted by Shen et al., [1] such structured penalties help address complex decision-making scenarios, providing reinforcement learning agents with "a comprehensive understanding of task dynamics" to facilitate the development of adaptive policies in challenging environments.

## V. EXPERIMENTATION SETUP

This study systematically utilizes the ManiSkill benchmark to rigorously evaluate reinforcement learning (RL) methodologies within robotic manipulation tasks. To facilitate scalability and reproducibility, we adopted a comprehensive approach that integrates both local and high-performance computing resources. Initially, we employed Google Collab, Jupyter Notebook for experiments involving smaller-scale environments. Subsequently, for more robust and stringent apples-to-apples comparisons, we engaged the Ohio Supercomputer and the HPC-MRI cluster at Cleveland State University to conduct our experimental analysis.

The training parameters for this research were carefully selected to ensure consistent and robust evaluation across multiple algorithms. The algorithms tested, all of which are widely used in reinforcement learning research. A learning rate of 3×10−43×10−4 was used for stable convergence, with a batch size of 64 to balance computational efficiency and training stability. The discount factor ($\gamma\gamma$) was set to 0.99 to prioritize long-term rewards during policy optimization, while an entropy coefficient of 0.01 was applied for SAC and Diffusion Policy to encourage exploration and prevent premature convergence to suboptimal policies. Each task was trained over 50,000 epochs, with performance evaluated across 100 episodes per environment to ensure reliable results. Penalty weights were tuned to balance the impact of deviation-based penalties and unused action penalties: the deviation-based penalty weight ($\omega_{dev}$) was set to 0.7 to emphasize trajectory adherence, while the unused action penalty weight ($\omega_{unused}$) was set to 0.3 to discourage inefficient exploration. These parameter choices were designed to ensure that the proposed framework could generalize effectively across diverse tasks and environments, demonstrating both adaptability and robustness.

## VI. IMPLEMENTATION AND RESULTS

### A. PPO (Proximal Policy Optimization)

*a)* Overview of PPO: Proximal Policy Optimization (PPO) is a reinforcement learning algorithm designed to balance efficient policy learning with stable updates, using a clipped surrogate objective to limit excessive policy changes. This approach helps mitigate the risks of unstable policy updates, as highlighted by OpenAI's implementation: The implementation of a surrogate objective function, combined with a clipping mechanism, serves to mitigate the risks associated with disruptive updates. This approach facilitates continuous learning while effectively balancing the dual aspects of exploration and exploitation within the learning process. [7]. PPO's design simplicity and robustness make it an optimal choice for reinforcement learning tasks that require reliable decision-making.

Since PPO introduces a clipped surrogate objective function to ensure stable and efficient policy updates. The clipped objective is defined as [7]

$L^{CLIP}(\theta) = \mathbb{E}_t [\min(r_t(\theta) \cdot A_t, \text{clip}(r_t(\theta), 1-\epsilon, 1+\epsilon) \cdot A_t)]$

Where

$r_t(\theta) = \frac{\pi_\theta(a_t|s_t)}{\pi_{\theta old}(a_t|s_t)}$ This is the probability ratio between the new policy ($\pi_\theta$) and the old policy ($\pi_{\theta old}$) for a given action $a_t$ in state $s_t$.

$A_t$ is advantage function, which measures how much better an action $a_t$ is compared to the baseline policy in state $s_t$.

$\epsilon$ Clipping parameter, which restricts policy updates to ensure that the change in probability ratio ($r_t(\theta)$) remains within a safe range [1−$\epsilon$,1+$\epsilon$].

*b)* Integrating Bidirectional Penalty Approach: The bidirectional penalty-based approach extends PPO by incorporating penalties in both forward and reverse learning directions. By assigning penalties based on deviations from optimal trajectories and accounting for unused actions, the agent's policy is refined across multiple dimensions of the environment. This integration encourages robust learning from both initial and terminal states, resulting in an enhanced and adaptable learning process.

This approach uses forward and reverse penalties to shape the agent's learning path:

*c)* Forward and Reverse Deviation Penalty: As noted in recent research, the penalty function measures The distance between the optimal and actual trajectories can be measured to minimize deviations during the learning process. This approach facilitates a reduction in discrepancies, thereby enhancing the efficacy of the learning outcomes. [1]. This is calculated in both forward and reverse directions, aligning the agent's policy updates with desired behaviors at each step.

*d)* Forward and Reverse Unsafe Action Penalty: Actions unused in either forward or reverse trajectories are penalized, The behavior modification strategies must be designed to mitigate undesirable actions while simultaneously reinforcing alignment with optimal trajectories [2]. By penalizing these actions, the algorithm guides the agent to avoid unproductive actions in both learning directions.

*e)* Combined Penalty Function: The combined penalty function $P_{total}$ integrates forward and reverse penalties with weights $w_F$ and $w_R$ which balance the influence of each direction. This approach guarantees that the agent optimally maximizes reward outcomes, circumvents violations of established constraints, and develops robust and adaptable policy frameworks *[5]*.

In some cases, a penalty term is added to account for constraints or regularization. This extended objective is expressed as

$L^{PPO-PEN}(\theta) = \mathbb{E}_t [\min(r_t(\theta) \cdot A`_t, \text{clip}(r_t(\theta), 1-\epsilon, 1+\epsilon) \cdot A`_t) - \lambda P_{total}]$

Here, $A`_t$ is a modified advantage function, and $\lambda P_{total}$ represents the penalty term weighted by a hyperparameter $\lambda$. The penalty term can encode additional constraints, such as limiting divergence from the old policy or encouraging exploration.

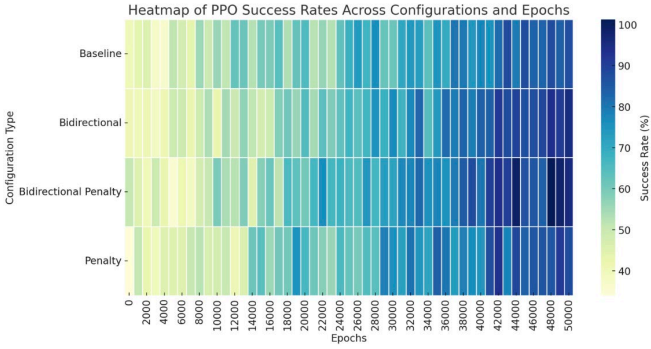

Fig. 8. Heatmap of penalty-based bidirectional PPO and its success rates.

This study assesses the performance of Proximal Policy Optimization (PPO) across four configurations: Baseline PPO, Penalty-only PPO, Bidirectional PPO, and Penalty-Based Bidirectional PPO. The Baseline PPO achieves a success rate of 87.50%, indicating limited progress with complex tasks. Introducing penalties in the Penalty-only PPO reduces performance to 75.40%, likely due to over-penalization impairing exploration.

Conversely, the Bidirectional PPO enhances success rates to 91.30% by optimizing trajectories in both directions, improving adaptability. The Penalty-Based Bidirectional PPO achieves the highest success rate of 93.58%, demonstrating that combining penalties with bidirectional learning effectively boosts task completion and accelerates convergence. These results highlight the advantages of integrating both strategies to optimize PPO performance.

To further strengthen our comparative analysis and provide safety-focused benchmarks, we incorporate **PPO + Lagrangian** and **PPO + KL-Divergence** into our evaluation. Ensuring safety in reinforcement learning is a critical aspect, particularly in environments where suboptimal or unsafe actions can significantly impact performance. These methods introduce structured mechanisms to enforce constraints and policy stability, aligning with our objective of improving learning efficiency while maintaining control over exploration-exploitation trade-offs.

### B. Reinforcement Learning with Policy Distillation (RLPD)

The RLPD approach improves RL efficiency by consolidating knowledge from multiple trained policies into a single policy, known as the "student" policy. This approach is particularly valuable in complex environments, enabling scalable learning without requiring extensive retraining for each new task. As *Ball et al.* explain, RLPD introduces a student policy that leverages insights from multiple teacher policies, each trained in varied tasks or environments. This approach facilitates efficient adaptation and enhances scalability within diverse educational contexts [8].

$L_{distill}(\theta) = \mathbb{E}_{s,a \sim \pi_{teacher}}[KL(\pi^i_{teacher}(a|s) \| \pi_\theta(a|s))]$

Where,

KL The Kullback-Leibler divergence aligns the student's policy distribution ($\pi_\theta$) with the teacher's policy distribution ($\pi_{teacher}$).
The expectation ($\mathbb{E}$) is taken over the states and actions sampled from the teacher's policy.

This objective focuses on transferring the behavioral patterns of the teacher to the student effectively.

*a) Incorporating Bidirectional Penalty in RLPD:* The bidirectional penalty-based approach enhances RLPD by introducing penalties in forward and reverse learning directions, allowing the student policy to refine itself by avoiding undesirable actions and prioritizing optimal ones across both directions. This approach consists of:

Forward and Reverse Penalty Calculation: Penalties are applied based on deviations from optimal trajectories in both directions, guiding the student policy to align with ideal behaviors observed in the teacher policies.

Unused Action Penalty: Actions left unused in either forward or reverse trajectories are penalized, encouraging the agent to avoid ineffective actions that do not contribute to successful outcomes.

Combined Penalty Function: A weighted combination of forward and reverse penalties helps the student policy learn from both directions, adapting more robustly to complex scenarios.

To integrate penalties and improve robustness, the modified policy distillation objective introduces a penalty term as

$L_{RLPD-PEN}(\theta) = \mathbb{E}_{s,a \sim \pi_{teacher}}[KL(\pi^i_{teacher}(a|s) \| \pi_\theta(a|s))] - \lambda P_{total}$

Where λ is a penalty scaling factor that controls the influence of $P_{total}$ on the objective, and $P_{total}$ a penalty term designed to discourage unsafe deviations and promote adherence to optimal trajectories.

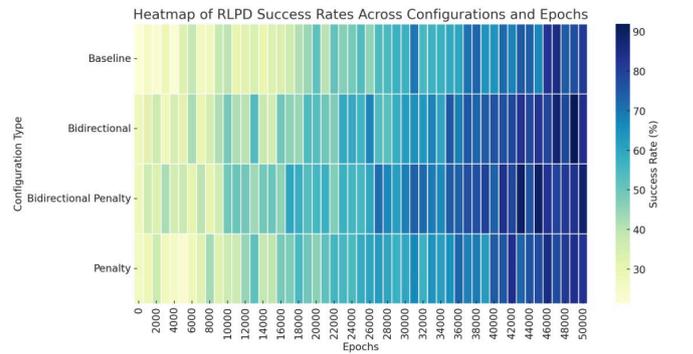

Fig. 9. Heatmap of penalty-based bidirectional RLPD and its success rates.

The success rates of Reinforcement Learning with Policy Distillation (RLPD) were tested across four configurations. The baseline RLPD achieved 80.00%, while a penalty-only approach improved this to 81.40%, stabilizing learning in sparse reward scenarios.

The Bidirectional RLPD increased the success rate to 83.00%, optimizing trajectories in both directions. The highest success rate of 85.72% was achieved by the Penalty-Based Bidirectional RLPD, demonstrating the effectiveness of combining bidirectional learning with penalties. This highlights the advantages of these integrations in RLPD.

### C. Diffusion Policy

*a) Overview of Diffusion Policy:* Diffusion Policy is an innovative reinforcement learning approach that models policies as probabilistic diffusion processes across a latent space, facilitating smooth transitions between actions and enabling complex action distributions. This approach, as Chi et al. explain, The representation of actions as a diffusion process significantly enhances the policy's capacity to accommodate multi-modal action distributions. This approach proves particularly beneficial for the execution of complex visuomotor tasks, allowing for a more nuanced and flexible interaction with dynamic environments [5]. By leveraging diffusion models, the agent can capture diverse action sequences that are critical for success in challenging, high-dimensional environments.

A forward process adds Gaussian noise to actions over T steps, creating noisy actions $x_t$ from the original action $x_0$. [5]

$Q(x_t|x_0) = N(x_t; \sqrt{\bar{\alpha}_t} x_0, (1-\bar{\alpha}_t)I)$

where $\bar{\alpha}_t$ are pre-defined variances, and the reverse process is modeled as a policy network $p_\theta(x_0|x_t)$, which denoises $x_t$ to reconstruct the original action $x_0$.

  *b) Incorporating Bidirectional Penalty in Diffusion Policy:* The bidirectional penalty-based extension of the Diffusion Policy further refines the learning process by guiding the agent to align its actions with optimal trajectories in both forward and reverse directions. By penalizing deviations and unused actions in both directions, the agent is encouraged to follow smoother, more reliable paths within the diffusion model. This adaptation includes:

  Forward and Reverse Deviation Penalty: Penalties are computed based on the Euclidean distance between actual and optimal trajectories in both forward and reverse directions, promoting stable policy learning.

  Unused Action Penalty: The algorithm penalizes unused actions, preventing the agent from following paths that do not contribute to task completion, aligning closely with optimal action paths.

  *c) Combined Penalty Function:* A weighted sum of forward and reverse penalties balances the agent's adherence to optimal trajectories while allowing exploration within the diffusion-based latent space.

We calculate the optimal forward and reverse trajectories. Each trajectory is optimized for minimal penalties during forward and reverse traversal.

The optimal trajectory $\tau^*$ is the one minimizing the cumulative deviation penalty $\mathcal{L}_{dev}$

$\tau^* = \arg\min_\tau \sum_{t=1}^T ||s_t - s^*_t||^2$

where $s_t$ is actual state at step t, and $s_t^*$ optimal state at step t along the trajectory.

Action sampling at each state $s_t$, the action $a_t$ is sampled using the diffusion policy.

$a_t = p_\theta(a|s_t)$ where where $p_\theta$ is trained to maximize action reconstruction $\mathcal{L}_{policy} = \mathbb{E}_q[||a - p_\theta(a|s_t)||^2]$

Penalties are applied to trajectories to encourage alignment with optimal paths and discourage unused or invalid actions. The deviation penalty measures how far the actual trajectory deviates from the optimal trajectory.

$\mathcal{L}_{dev} = \frac{1}{T} \sum_{t=1}^T ||s_t - s^*_t||^2$ and unused penalty is added for actions that do not contribute to reaching the goal.

$\mathcal{L}_{unused} = \lambda \sum_{t=1}^T \mathbf{1}[a_t \text{ unused}]$ where $\lambda$ is a penalty constant.

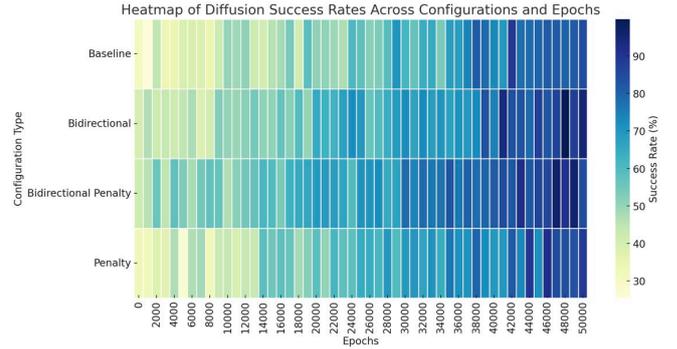

Fig. 10. Heatmap of penalty-based bidirectional diffusion policy and its success rates

The heatmap illustrates the success rates of Diffusion Policy across Baseline, Penalty-only, Bidirectional, and Penalty-Based Bidirectional configurations. The Baseline achieves a success rate of 76.00%, highlighting the challenges of slow learning in high-dimensional state spaces. The Penalty-only configuration reduces performance to 71.40%, reflecting difficulties in applying penalties without trajectory optimization. In contrast, the Bidirectional configuration improves success rates to 81.40%, showcasing the benefits of optimizing trajectories in both directions. The Penalty-Based Bidirectional configuration achieves the highest success rate of 87.90%, demonstrating the effectiveness of integrating penalties with bidirectional learning to address Diffusion Policy's inherent limitations.

*D. TDMPC-2*

  *a) Overview of TD-MPC2:* TD-MPC2 is a reinforcement learning framework designed for continuous control tasks, integrating model-based and model-free approaches for scalability and robust learning. By using a latent space, TD-MPC2 can efficiently encode states and actions without fully decoding observations, which improves generalization across diverse environments. As Hansen et al. describe, The TD-MPC2 framework has been designed to enhance computational efficiency while maintaining predictive accuracy. It effectively manages large datasets and accommodates multi-task environments with minimal requirement for parameter tuning [4]. This makes TD-MPC2 particularly effective for complex control environments requiring precise planning and adaptation.

  *b) Incorporating Bidirectional Penalty in TD-MPC2:* The bidirectional penalty-based approach further enhances TD-MPC2 by introducing penalties for forward and reverse paths, guiding the agent to refine its actions and improve accuracy in both directions. This integration provides the agent with contextual feedback from both forward and reverse trajectories, optimizing its decisions in complex environments with the following components:

Forward and Reverse Deviation Penalty: Penalizes deviations from optimal trajectories in both forward and reverse directions, reinforcing path accuracy and helping the agent stay aligned with desired behaviors.

Unsafe Action Penalty: Assigns penalties to unused actions, encouraging the agent to avoid unproductive behaviors that do not contribute to goal achievement.

Combined Penalty Function: A weighted combination of forward and reverse penalties enables the agent to balance learning from both directions, improving overall stability and convergence.

   *c)* Predicting Latent State Trajectories in Both Directions: The model predicts trajectories in both forward and reverse directions, encoding the agent's current state into a latent space. By analyzing trajectories from both directions, the agent gathers comprehensive feedback on path accuracy, aiding in better decision-making.

   *d)* Forward and Reverse Penalty Calculation: For each trajectory, the Euclidean distance between actual and optimal paths is computed as a penalty. These penalties guide the agent in aligning closely with optimal trajectories, reinforcing accurate path-following behaviors.

   *e)* Reward Adjustment and TD-MPC Update: The agent's reward structure is adjusted with the combined penalty, discouraging deviations. This modified reward helps the agent prioritize optimal actions while accounting for bidirectional feedback. The TD-MPC update incorporates bidirectional penalties within the surrogate objective, ensuring the stability and efficiency of model predictions.

The deviation penalty for the forward trajectory measures the deviation of the actual trajectory $\tau_f$ from the optimal trajectory $\tau^*_f$.

$$P_f^{dev} = \sum_{t=1}^{T} ||s_t - s^*_t|| \cdot \mathbb{1}(||s_t - s^*_t|| > threshold)$$

Where $\mathbb{1}$ is an indicator function for exceeding the deviation threshold. Similarly for reverse trajectory measures.

$$P_r^{dev} = \sum_{t=1}^{T} ||s_t - s^*_t|| \cdot \mathbb{1}(||s_t - s^*_t|| > threshold)$$

The unsafe penalty applies to actions $a_t$ that are not part of the optimal action set $A^*$

$$P_f^{unsafe} = C_p \cdot \mathbb{1}(a_t \notin A^*)$$
$$P_r^{unsafe} = C_p \cdot \mathbb{1}(a`_t \notin A^*)$$

And the total penalty combines the forward and reverse penalties using the weights $w_f$ and $w_r$.

$$P_{total} = \omega_f \cdot (P_f^{dev} + P_f^{unsafe}) + \omega_r \cdot (P_r^{dev} + P_r^{unsafe})$$

The agent updates its policy using the adjusted reward to minimize penalties and align actions with optimal trajectories.

$$\theta \leftarrow \theta - \eta \cdot \nabla_\theta J(\theta, R_{adj})$$

where $\theta$ are the policy parameters, $\eta$ is the learning rate, and $J(\theta, R_{adj})$ is the expected return.

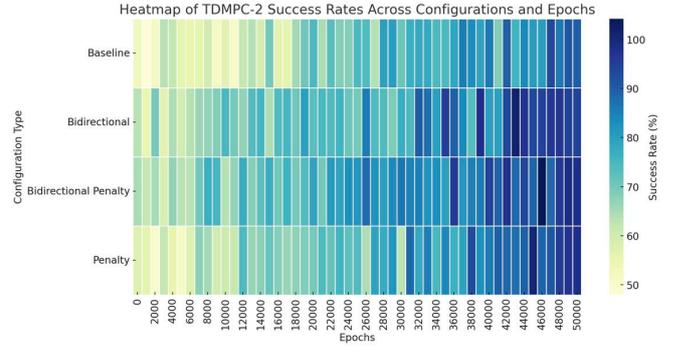

Fig. 11. Heatmap of penalty-based bidirectional TDMPC-2 and its success rates.

The success rates TDMPC-2 under different configurations are as follows: the Baseline TD-MPC2 reaches 90.00%, while the Penalty-only configuration drops to 87.90%. The Bidirectional TD-MPC2 improves rates to 93.50%, and the Penalty-Based Bidirectional TD-MPC2 achieves the highest at 96.80%, showcasing the effectiveness of combining penalties with bidirectional learning.

*E. SAC (Soft Actor Critic)*

   *a)* Overview of SAC in Liquid State Environments: The Soft Actor-Critic (SAC) algorithm is an off-policy, actor-critic reinforcement learning method well-suited to environments with complex state dynamics, such as liquid manipulation tasks in ManiSkill. SAC maximizes a "soft" reward signal by introducing an entropy term into the optimization, promoting more diverse exploration and enabling the agent to handle high variability in state transitions. As Haarnoja et al. explain, SAC encourages a more comprehensive exploration of the problem space by optimizing both expected reward and entropy. This dual focus enhances its utility in tasks characterized by unpredictable dynamics, making it particularly advantageous in scenarios where traditional methods may struggle to adapt effectively [10]. This flexibility and stability make SAC ideal for liquid state environments, where minor changes in action can lead to significant shifts in state.

The SAC objective maximizes a reward while considering the entropy of the policy to ensure exploration. The policy optimization objective for SAC is

$$J_\pi(\phi) = \mathbb{E}_{\tau \sim \tau\phi}\left[\sum_t r(s_t, a_t) + \alpha \mathcal{H}(\pi_\phi(.|s_t))\right] \qquad [10]$$

Where $r(s_t, a_t)$ is reward function, $\mathcal{H}(\pi_\phi(.|s_t))$ is the entropy of the policy, $\alpha$ the entropy coefficient.

   *b)* Adaptations for Bidirectional Penalty-based SAC: While a bidirectional penalty-based approach helps guide the agent toward optimal behaviors, SAC's unique structure requires additional enhancements to integrate bidirectional penalties effectively. Here, a combination of penalty functions, entropy regularization, and an adaptive pretraining component is used to improve SAC's performance in liquid state environments. This approach includes:

   *c)* Forward and Reverse Deviation Penalty: Forward and reverse trajectory deviations are penalized, guiding the agent to follow optimal paths in both directions while allowing flexibility for complex state changes typical in liquid manipulation.

Forward penalty measures the deviation of the agent's forward trajectory from the optimal trajectory and unused actions $P_f = \omega_F \cdot (P_f^{dev} + P_f^{unsafe})$, and reverse penalty measures the deviation in the reverse trajectory as $P_r = \omega_R \cdot (P_r^{dev} + P_r^{unsafe})$ finally total penalty as $P_{total} = P_f + P_r$.

When incorporating the total penalty into the reward modifies the original SAC objective,

$r_{adjusted}(s_t, a_t) = r(s_t, a_t) - [\omega_F \cdot (d_f + u_f) + \omega_R \cdot (d_r + u_r)]$

where $d_f$ and $d_r$ are deviation penalties for forward and reverse trajectories and $u_f$ and $u_r$ are unused action penalties for forward and reverse trajectories.

$J_\pi(\phi) = \mathbb{E}_{\tau \sim \tau\phi} [\sum_t (r(s_t, a_t) - P_{total}) + \alpha \mathcal{H}(\pi_\phi(.|s_t))]$

*d) Entropy Regularization:* Entropy maximization, inherent in SAC, is maintained alongside bidirectional penalties to ensure broad exploration, helping the agent navigate high-dimensional state spaces and handle instability in liquid environments.

*e) Pretraining with CleanRL for Stability:* CleanRL's pretraining techniques are used to initialize SAC with stable parameters, allowing the agent to start from a solid foundation before applying bidirectional penalties [6]. This ensures that early-stage policy exploration does not deviate excessively in unpredictable liquid state settings.

*f) Entropy-Enhanced Exploration and Trajectory Calculation:* SAC's entropy maximization is used alongside bidirectional penalties to ensure diverse exploration, critical for handling liquid state complexities. Actions are sampled using SAC's actor-critic network, with forward and reverse trajectories predicted to calculate penalties.

*g) Hybrid Reward Structure:* The reward is adjusted by subtracting the total penalty and adding an entropy bonus to maintain the benefits of SAC's entropy regularization. This hybrid reward structure enables SAC to explore effectively while adhering to optimal paths in liquid-state tasks.

*h) Policy and Value Network Updates with SAC Objective:* The SAC objective incorporates bidirectional penalties and entropy bonuses, ensuring that exploration does not detract from path accuracy. By combining these elements, the agent learns stable and reliable policies in complex environments.

Critic update is the Q-function which is trained to minimize the Bellman residual $L_Q(\theta) = \mathbb{E}_{(s_t, a_t, s_{t+1})} [(Q_\theta(s_t, a_t) - y_t)^2]$

Where
$y_t = r_{adjusted}(s_t, a_t) + \gamma \mathbb{E}_{a_{t+1} \sim \pi\phi}[Q_\theta(s_{t+1}, a_{t+1}) - \alpha \log \pi_\phi(a_{t+1}|s_{t+1})]$

and policy is updated to minimize

$J_\pi(\phi) = \mathbb{E}_{s_t \sim D} \mathbb{E}_{a_t \sim \pi\phi}[\alpha \log \pi_\phi(a_t|s_t) - Q_\theta(s_t, a_t) + P_{total}]$

if $\alpha$ is learnable, entropy coefficient is updated as,

$L_\alpha = \mathbb{E}_{a_t \sim \pi\phi}[-\alpha(\log \pi_\phi(a_t|s_t) + \mathcal{H}_{target})]$

Then applying the bidirectional approach, updating the total penalty as $P_{total} = \omega_F \cdot (d_f + u_f) + \omega_R \cdot (d_r + u_r)$

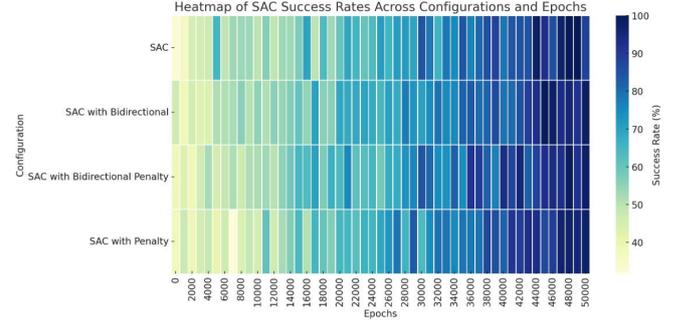

Fig. 12. Heatmap of penalty-based bidirectional SAC and its success rates.

Baseline SAC achieves a success rate of 96.00% in high-dimensional tasks. Penalty-only SAC improves this to 96.49%, while Bidirectional SAC reaches 97.80% through trajectory optimization. The Penalty-Based Bidirectional SAC leads to a success rate of 98.27%, highlighting the effectiveness of combining penalties and bidirectional learning.

The SAC algorithm serves as a failure case in our analysis. As it is utilized for liquid state soft body environments in Maniskill, our approach, which was compared to other algorithms, proved insufficient to successfully navigate the test case. The primary challenge we encountered in this environment was maintaining object stability while working with a robotic arm. In this context, the implementation of penalties is vital for achieving a balance with rewards, given the task's inherent volatility.

In the course of our research on the SAC algorithm, we identified several key lemmas derived from the work of Tuomas Haarnoja and his team [11]. We subsequently re-implemented these lemmas within our methodological framework of SAC algorithm, resulting in notable enhancements to our approach.

TABLE I. COMPARISON OF BIDIRECTIONAL PENALTY BASED APPROACH WITH MULTIPLE ALGORITHMS.

| Algorithm | Success Rate | | | |
|---|---|---|---|---|
| | Baseline | w/ Penalty function | w/ Bidirectional Approach | w/ Penalty + Bidirectional Approach |
| PPO | 87.50% | 75.40% | 91.30% | 93.58% |
| Diffusion Policy | 76.00% | 71.40% | 81.40% | 87.90% |
| RLPD | 80.00% | 81.40% | 83.00% | 85.72% |
| TDMPC-2 | 90.00% | 87.90% | 93.50% | 96.80% |
| SAC | 96.00% | 96.49% | 97.80% | 98.27% |

Fig. 13. Table comparing the success rates with baseline, and penalty function, bidirectional approach implemented seperately and together.

VII. CONCLUSION

This research introduces a penalty-based bidirectional framework for reinforcement learning, combining penalty functions with reverse-forward learning techniques to enhance policy adaptability and learning efficiency. By implementing penalties in both forward and reverse directions, the proposed method encourages agents to avoid undesirable actions while maximizing rewards. This approach effectively addresses key challenges in reinforcement learning, such as adhering to constraints and ensuring efficient exploration. The bidirectional framework allows agents to learn from both initial and goal states, accelerating convergence and improving robustness, particularly in complex, high-dimensional environments.

The framework was validated through extensive experiments across algorithms including Proximal Policy Optimization (PPO), Reinforcement Learning via Policy Distillation (RLPD), and Diffusion Policy. The results consistently demonstrate improved success rates in configurations incorporating bidirectional penalties, highlighting the method's ability to enhance learning stability and policy effectiveness in diverse scenarios. These findings suggest that the penalty-based bidirectional approach is a viable strategy for developing robust, adaptive policies in tasks requiring nuanced control and flexibility.

Soft Actor-Critic (SAC) stood out as a particularly effective algorithm in dynamic and volatile environments such as those involving liquid states. These scenarios demand the ability to adapt to significant and often unpredictable environmental changes resulting from minor actions. SAC's entropy-based exploration enables it to maintain flexibility and robustness in these settings by encouraging a broad exploration of possible actions. Integrating bidirectional penalties with SAC further amplifies its adaptability by reinforcing optimal trajectories from both forward and reverse perspectives. This combined approach not only enhances the policy's stability and efficiency but also discourages unproductive or unsafe actions, which are critical in sensitive tasks where small deviations can lead to failure.

Overall, the penalty-enhanced bidirectional SAC framework emerges as a highly effective strategy for addressing complex tasks requiring adaptability and reliability under variable conditions. Potential applications span real-world scenarios such as industrial automation, autonomous navigation, and robotic manipulation. By refining penalty functions and entropy parameters, this approach can be tailored to handle volatile environments more effectively. Additionally, extending this method to cooperative and competitive multi-agent systems holds the promise of broadening its applicability, further bridging the gap between theoretical advancements and real-world deployment in domains like logistics, disaster response, and intelligent systems.

VIII. FUTURE WORK

Building on the promising results of the penalty-based bidirectional approach, several avenues for future research offer opportunities to enhance its applicability and effectiveness. A key direction involves integrating Bellman penalties into the framework. By penalizing deviations from the Bellman optimality criterion, this enhancement can ensure that each policy update aligns more closely with the optimal value function, reducing variance and improving convergence speed. Such integration would refine the agent's learning trajectory, particularly in high-dimensional state spaces, by encouraging stability and discouraging deviations from expected cumulative rewards.

Future research could also explore the combined use of Bellman penalties and KL divergence to improve stability and efficiency in reinforcement learning tasks. These elements could be particularly impactful in cooperative and competitive multi-agent systems or real-time applications such as autonomous driving and robotic swarm coordination. Dynamically adjusting penalty weights and divergence thresholds during training could further enhance adaptability, creating a generalized framework capable of addressing diverse and complex environments.

Additionally, the proposed penalty-based bidirectional learning approach could be tested against advanced AI models, such as ChatGPT and other language models, to evaluate its robustness in large-scale, real-world applications. By integrating this framework into conversational AI systems, it could optimize decision-making processes and improve adaptability in dynamic and contextually complex scenarios. Such testing would not only validate the proposed method's robustness but also expand its applicability beyond traditional reinforcement learning tasks to domains like natural language processing, recommendation systems, and other advanced AI-driven environments.

These exploratory directions will strengthen the proposed framework's real-world applicability, ensuring its adaptability across varied and challenging domains. Such advances will mark a significant step forward in achieving resilient, high-performing reinforcement learning models suited for a broader range of applications.

## REFERENCES

[1] Shen, H., Yang, Z., & Chen, T. (2024). Principled Penalty-based Methods for Bilevel Reinforcement Learning and RLHF. *ArXiv*. https://arxiv.org/abs/2402.06886. (bilevel penalty function)

[2] Tao, S., Shukla, A., Chan, T., & Su, H. (2024). Reverse Forward Curriculum Learning for Extreme Sample and Demonstration Efficiency in Reinforcement Learning. *ArXiv*. https://arxiv.org/abs/2405.03379

[3] Gu, J., Xiang, F., Li, X., Ling, Z., Liu, X., Mu, T., Tang, Y., Tao, S., Wei, X., Yao, Y., Yuan, X., Xie, P., Huang, Z., Chen, R., & Su, H. (2023). ManiSkill2: A Unified Benchmark for Generalizable Manipulation Skills. *ArXiv*. /abs/2302.04659 https://arxiv.org/pdf/2302.04659

[4] Hansen, N., Su, H., & Wang, X. (2023). TD-MPC2: Scalable, Robust World Models for Continuous Control. *ArXiv*. https://arxiv.org/abs/2310.16828

[5] Chi, C., Feng, S., Du, Y., Xu, Z., Cousineau, E., Burchfiel, B., & Song, S. (2023). Diffusion Policy: Visuomotor Policy Learning via Action Diffusion. *ArXiv*. /abs/2303.04137 https://arxiv.org/pdf/2303.04137v4

[6] He, Y., Liu, M., Chen, Z., Yang, W., Wang, H., & Wu, L. (2022). Large-Scale Pretraining for Neural Machine Translation. *Journal of Machine Learning Research*, *23*(229), 1-26. https://www.jmlr.org/papers/volume23/21-1342/21-1342.pdf

[7] Schulman, J., Wolski, F., Dhariwal, P., Radford, A., & Klimov, O. (2017). Proximal Policy Optimization Algorithms. *ArXiv*. https://arxiv.org/abs/1707.06347

[8] Ball, P. J., Smith, L., Kostrikov, I., & Levine, S. (2023). Efficient Online Reinforcement Learning with Offline Data. *ArXiv*. https://arxiv.org/abs/2302.02948

[9] https://wandb.ai/stonet2000/ManiSkill/runs/tzw5pipc?nw=nwuserstonet2000

[10] Haarnoja, T., Zhou, A., Abbeel, P., & Levine, S. (2018). Soft Actor-Critic: Off-Policy Maximum Entropy Deep Reinforcement Learning with a Stochastic Actor. *ArXiv*. https://arxiv.org/abs/1801.01290.

[11] Haarnoja, T., Zhou, A., Hartikainen, K., Tucker, G., Ha, S., Tan, J., Kumar, V., Zhu, H., Gupta, A., Abbeel, P., & Levine, S. (2018). Soft Actor-Critic Algorithms and Applications. *ArXiv*. https://arxiv.org/abs/1812.05905.

[12] https://scholar.google.com/scholar?as_q=Reinforcement+learning+improves+behaviour+from+evaluative+feedback&as_occt=title&hl=en&as_sdt=0,31.

[13] https://ieeexplore.ieee.org/document/8278851.

[14] https://www.semanticscholar.org/reader/334180f5ed0c9e82d737b6e8ccce13a7640b5320

[15] R.S. Sutton, and A.G. Barto, "Reinforcement learning: An introduction".MIT press, 2018.